\documentclass[conference]{IEEEtran}
\IEEEoverridecommandlockouts
\usepackage{multirow}
\usepackage{cite}
\usepackage{amsmath,amssymb,amsfonts,algpseudocode}
\usepackage{algorithm}
\usepackage{graphicx}
\usepackage{textcomp}
\usepackage{xcolor}
\usepackage{upgreek}
\usepackage{optidef}
\usepackage{breqn}
\usepackage{varwidth}
\usepackage{booktabs}
\usepackage{subcaption}
\usepackage[utf8]{inputenc}

\def\BibTeX{{\rm B\kern-.05em{\sc i\kern-.025em b}\kern-.08em
    T\kern-.1667em\lower.7ex\hbox{E}\kern-.125emX}}
\begin{document}

\title{Robust Federated Learning on Edge Devices with Domain Heterogeneity\\

}

\author{\IEEEauthorblockN{Huy Q. Le\IEEEauthorrefmark{2}, Latif U. Khan\IEEEauthorrefmark{3} and Choong Seon Hong\IEEEauthorrefmark{2}}
\IEEEauthorblockA{\IEEEauthorrefmark{2}Department of Computer Science and Engineering, Kyung Hee University, 171-04, Republic of Korea}
\IEEEauthorblockA{\IEEEauthorrefmark{3}Machine learning department, Mohamed Bin Zayed University of Artificial Intelligence, United Arab Emirates}
Email:\{quanghuy69, cshong\}@khu.ac.kr, latif.u.khan2@gmail.com}
\maketitle

\begin{abstract}
Federated Learning (FL) allows collaborative training while ensuring data privacy across distributed edge devices, making it a popular solution for privacy-sensitive applications. However, FL faces significant challenges due to statistical heterogeneity, particularly domain heterogeneity, which impedes the global mode's convergence. In this study, we introduce a new framework to address this challenge by improving the generalization ability of the FL global model under domain heterogeneity, using prototype augmentation. Specifically, we introduce FedAPC (Federated Augmented Prototype Contrastive Learning), a prototype-based FL framework designed to enhance feature diversity and model robustness. FedAPC leverages prototypes derived from the mean features of augmented data to capture richer representations. By aligning local features with global prototypes, we enable the model to learn meaningful semantic features while reducing overfitting to any specific domain.  Experimental results on the Office-10 and Digits datasets illustrate that our framework outperforms SOTA baselines, demonstrating superior performance.
\end{abstract}


\section{Introduction}
The boomming era of artificial intelligence (AI) has transformed various fields, such as healthcare, recommendation systems~\cite{antunes2022federated,yang2020federated,banabilah2022federated,qiao2025deepseek}. These advancements rely on the personal data at the edge, raising the privacy concerns in centralized machine learning systems. Traditional centralized approaches require data to be collected and processed on a central server, which exposes the privacy leakage of user data. This challenge has urged the demand of achieving the edge intelligence in the decentralized manner. Federated Learning (FL) has gained recognition as an effective solution with the decentralized paradigm~\cite{mcmahan2016communication,kairouz2021advances} to address the above issue. Unlike conventional machine learning, federated learning (FL) enables edge devices to train a shared global model while keeping their private data secure. Instead, only model updates (e.g., gradients or paramaters) are communicated between devices and a server, thus preserving data privacy and reducing communication overhead.

\begin{figure}[]
    \centering
	\includegraphics[width=0.9\linewidth]{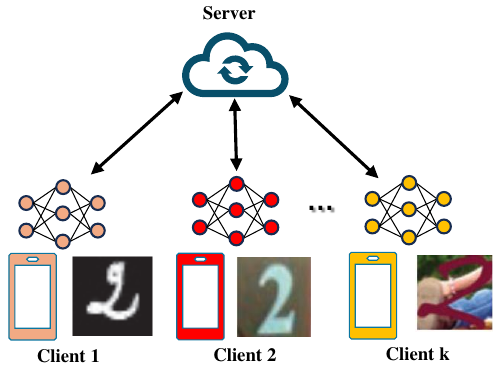}
	\caption{Problem illustration of federated learning under domain heterogeneity.}
	\label{illustration} 
\end{figure}

Although federated learning offers numerous advantages, it also presents significant challenges, particularly when dealing data heterogeneity (e.g, non-i.i.d data)~\cite{li2020federated,ye2023heterogeneous}. One of the key challenges resulting from non-IID data is feature heterogeneity, which refers to the differences in the feature distributions across various clients. For instance, in a visual recognition task, digit images collected from different clients may belong to the same category but originate from distinct domains, resulting in a diverse feature distribution across clients, as shown in Fig.~\ref{illustration}. The variation in data distributions across different clients results in a misalignment of feature spaces, which impedes the global model performance.
To address the challenge of domain heterogeneity in FL, federated prototype learning has emerged as an effective solution. By using prototypes, this method enables the representation of complex features in a compact, low-dimensional space, effectively reducing noise and outliers. This results in more robust and stable feature representations, which can significantly improve model performance. Prototype learning has been applied in various domains, including few-shot learning~\cite{zhang2021prototype} and unsupervised learning~\cite{song2024morphological}. In the context of FL, several methods have been proposed that leverage prototypes to improve model performance under data heterogeneity. FedProto~\cite{tan2022fedproto} was one of the first approaches to construct global prototypes from local features and then use these global prototypes as a form of regularization to align local features, thus addressing label heterogeneity. Recently, the authors in~\cite{le2024cross} proposed the concept of a complete prototype to tackle the challenge of missing modalities in multimodal federated learning. In addition, to mitigate the effects of domain shifts, various works~\cite{huang2023rethinking,qiao2025fedccl} have explored clustering-based methods to create unbiased prototypes that can facilitate better alignment during local training. These clustering approaches aim to create prototypes that better represent the global data distribution, reducing domain-specific biases. However, many of the existing federated learning methods that utilize prototypes primarily rely on prototypes derived from local data. While this approach works in many scenarios, it limits the diversity of the local feature representations within the same domain. 

In this work, we propose a novel solution to address the domain shift problem in FL by introducing the Federated Augmented Prototype Contrastive Learning (FedAPC) framework. To begin, we leverage multi-view feature representations to generate augmented prototypes, which capture a more comprehensive and diverse set of features from the data. By incorporating these augmented prototypes, we aim to enrich the feature space, facilitating better generalization across clients with varying data distributions. Next, we perform contrastive alignment between the local features and the global prototypes. This alignment helps ensure that the local features are closely aligned with the corresponding global prototypes, promoting consistency across different clients while reducing the impact of domain shifts. Through this process, we enhance the diversity of features within the same domain, thereby improving the robustness of the global model under domain heterogeneity. Our contributions in this paper are as follows:
\begin{itemize}
    \item We introduce a novel FL framework, FedAPC, designed to address domain heterogeneity in federated learning.
	\item We enhance the robustness of the global model under domain heterogeneity by leveraging augmented prototypes and performing contrastive alignment between the global prototypes and local features. 
	\item We evaluate the effectiveness of our approach on two common datasets: Digits and Office-10. The experimental results demonstrate faster convergence and improved average performance across different domains, outperforming existing baselines. 
\end{itemize}

\begin{figure*}[t]
	\centering
	\includegraphics[width=0.9\linewidth]{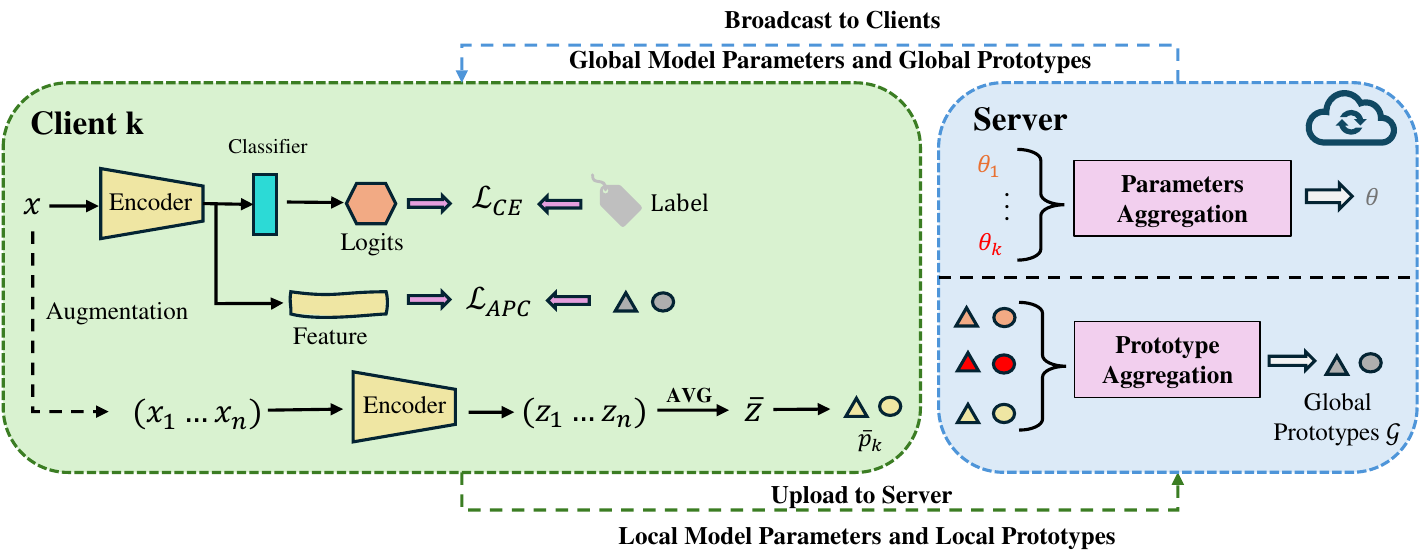}
	\caption{Illustration of FedAPC.}
	\label{systemmodel}
\end{figure*}

\section{Methodology}
\subsection{Preliminary}
 Our proposed framework is depicted in Fig.~\ref{systemmodel}. We consider a FL scenario with $K$ clients, where each client owns the private data $D_k=\{x_k,y_k\}$, with $x_k$ representing input samples and $y_k$ are the corresponding labels.  Specifically, in the context of domain shift, clients may possess different feature distributions, meaning that the feature distributions from client $i$ and client $j$ may not match,, i.e., $P^i_k(x)\ne P^j_k(x)$, but they share the same label space $P^i_k(y) = P^j_k(y)$.

 The global objective in is to train a global model $\theta$ that can generalize across different, despite the differences in their local data distributions. The global objective is defined as:
\begin{align}
     \mathcal{L}(\theta) = \frac{1}{K} \sum_{k=1}^{K} \mathbb{E}_{(x_k, y_k) \sim D_k} \mathcal{L}_k(\theta, x_k, y_k),
\end{align}
where $\mathcal{L}_k(\theta, x_k, y_k)$ is the local loss function for client $k$, and $D_k$ represents the data distribution of client $k$. 
\subsection{Proposed FedAPC framework}

\noindent\textbf{Local Augmented Prototypes.} All clients in the federated learning framework use the same model architecture, which includes a feature encoder $h$ and a classifier $\phi$. The feature extractor encodes the input sample $x$ into a feature vector $z=h(x)$. To enhance the feature diversity and robustness under domain heterogeneity, we employ data augmentation. For a given augmented data $x_n$, the augmented feature is denoted as $z_n=f(\mathcal{A}(x_n))$, where $\mathcal{A}(x_n)$ represents the augmented version of the input $x_n$ through transformation function $\mathcal{A}$. This augmentation helps to create multiple views of data, which improves the model's generalization across various domains. The mean feature across different augmented versions of the input data is denoted as follows:
\begin{align}
\bar{z} = \frac{1}{N}\sum_{n\in N} z_n,
\end{align}
where $N$ is the total number of augmented versions for each input. The mean feature $\bar{z}$ serves as a representative feature for the augmented data. Next, we define the augmented prototypes for each client as. We denote the local augmented prototypes as follows:
\begin{align}
    \bar{p}^m_{k}=\frac{1}{|S^m_{k}|}\sum_{i\in S^k_{m}}\bar{z}^i,
    \label{eq:localproto}
\end{align}
where $S^m_{k}$ refers to the subset of the client's private data $D_k$ corresponding to the $m-th$ class, and $\bar{z}^i$ denotes the mean feature for each data point $i$ of class $m$. The local augmented prototypes are then transmitted to the server.

\noindent\textbf{Global prototypes.} At the server, prototype aggregation is performed to compute the global prototypes, which are derived by averaging local augmented prototypes from all clients:
\begin{align}
    \begin{split}
    \mathcal{G}^m&=\sum^K_{k=1}\bar{p}^m_k\in\mathbb{R}^d 
    \\
    \mathcal{G}&=[\mathcal{G}^1,\mathcal{G}^2,\dots,\mathcal{G}^m],
    \end{split}
\label{global_proto}
\end{align}
Here, $\mathcal{G}^m$ represents the global prototypes for class $m$. 

\noindent\textbf{Augmented Prototypes Contrastive Learning.} These global prototypes are then broadcast to each client for prototype alignment during local training.
To facilitate this, we propose a prototype contrastive loss that encourages local features to align with global prototypes of the same semantic class while pushing away prototypes from different classes. The augmented prototypes contrastive loss is defined as:
\begin{align}
        \mathcal{L}_{APC} = -\log\frac{\sum_{g^+\in \mathcal{G}^+}\exp(s(h(x),g^+)/\tau)}{\sum_{g^k\in G}\exp(s(h(x),g^k)/\tau)}, 
       \label{loss_apc}
\end{align}
where $\mathcal{G}^+$ represents the global prototypes that share the same semantic class as the local features (e.g., positive class), $(s(h(x),g^+)$ denotes the cosine similarity function between the local feature $h(x)$ and prototype $g^+$, and $\tau$ is the temperature parameter. We compute the overall loss for each client as follows:
\begin{align}
    \mathcal{L}=\mathcal{L}_{CE} + \mathcal{L}_{APC},
    \label{overall}
\end{align}
where $\mathcal{L}_{CE}$ is the cross-entropy loss calculated between the predicted output and its corresponding label.

\subsection{FedAPC Algorithm}
This section presents the FedAPC algorithm, as outlined in Alg.~\ref{alg:FedAPC}. In each communication round, server broadcast the global model $\theta_t$ and global prototypes $\mathcal{G}$ to all clients. Each client then conducts the local training process with its private data $D_k$, updating the local models based on the overall training loss in Eq.~\ref{overall}. Once the local training is completed, clients extract their local augmented prototypes and transmit both the local prototypes and updated model parameters to the global server. Then the server performs aggregation, combining the received models and prototypes to update the global model and prototypes. This iterative process continues until the model reaches convergence.

\begin{algorithm}[t] 
    \caption{FedAPC} 
    \label{alg:FedAPC} 
    \begin{algorithmic}[1]
        \State \textbf{Input:}~~Dataset $D_k$ of each client.
        \State \textbf{Initialize $\theta^0$}
        \For{ $t$ = 1, 2, ..., $T$} 
            \For{ $k$ = 0, 1,..., $K$ \textbf{in parallel}}
                \State Broadcast global model $\theta^t$ to client \textit{k}
                \State {$\theta_k^t,\bar{p}^m_k \gets \textbf{LocalTraining}(\theta^t$, $\mathcal{G}$)}
            \EndFor
            \State Global prototype aggregation by Eq. \ref{global_proto}
            \State {$\theta^{t+1} \gets \sum_{k=1}^{K} \frac {D_k}{\sum_{k=1}^{K} D_k}\theta_k^t$}
        \EndFor \\
        \textbf {LocalTraining($\theta^t$, $\mathcal{G}$)}
        \For{ each local round }
        \For{ each batch ($\boldsymbol{x}_k$, $y_k$) of $D_k$ }
        \State {Local augmented prototypes are computed in Eq.~\ref{eq:localproto}} 
        \State $\mathcal L_{CE} \gets CrossEntropyLoss(\phi(z), y_k))$
        \State Calculate $\mathcal L_{APC}$ by Eq. \ref{loss_apc}
        \State 
$\mathcal{L}=\mathcal{L}_{CE}+\mathcal{L}_{APC}$ 
        \State {$\theta_k^t \gets \theta_k^t - \eta\nabla\mathcal L$}
    \EndFor
    \EndFor
        \State\textbf{Return:} $\theta_k^t$, $\bar{p}^m_k$
    \end{algorithmic}
\end{algorithm}
\begin{figure*}[]
    \centering
    \begin{subfigure}{0.40\textwidth}
        \centering
        \includegraphics[width=\linewidth]{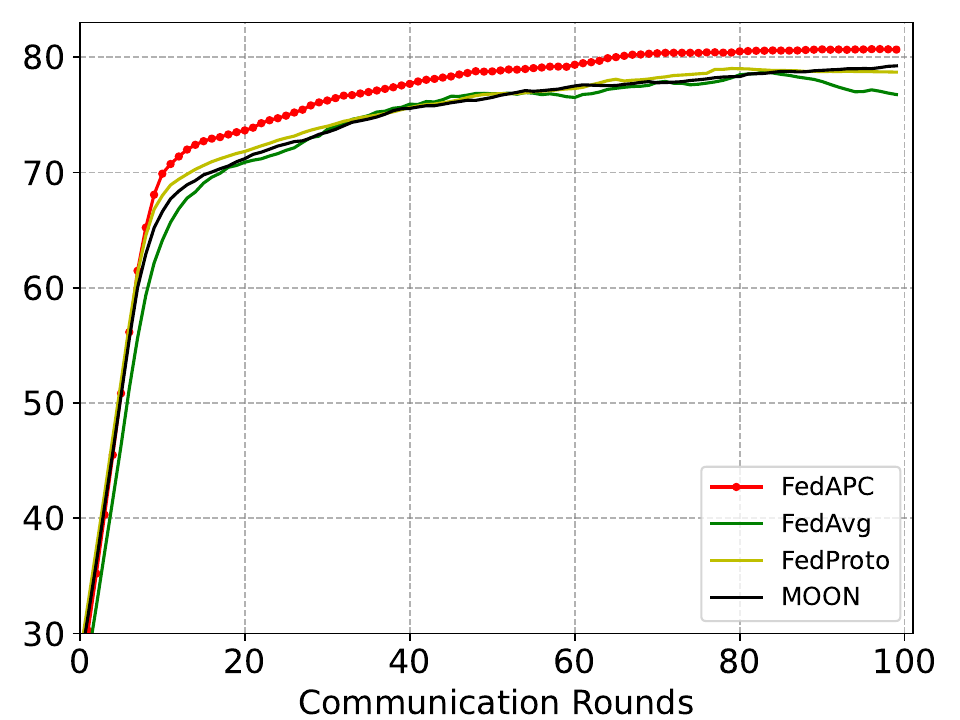}
        \caption{Digits}
        \label{fig:temperature_ucihar}
    \end{subfigure}%
\hspace{-0.05cm} 
    \begin{subfigure}{0.40\textwidth}
        \centering
        \includegraphics[width=\linewidth]{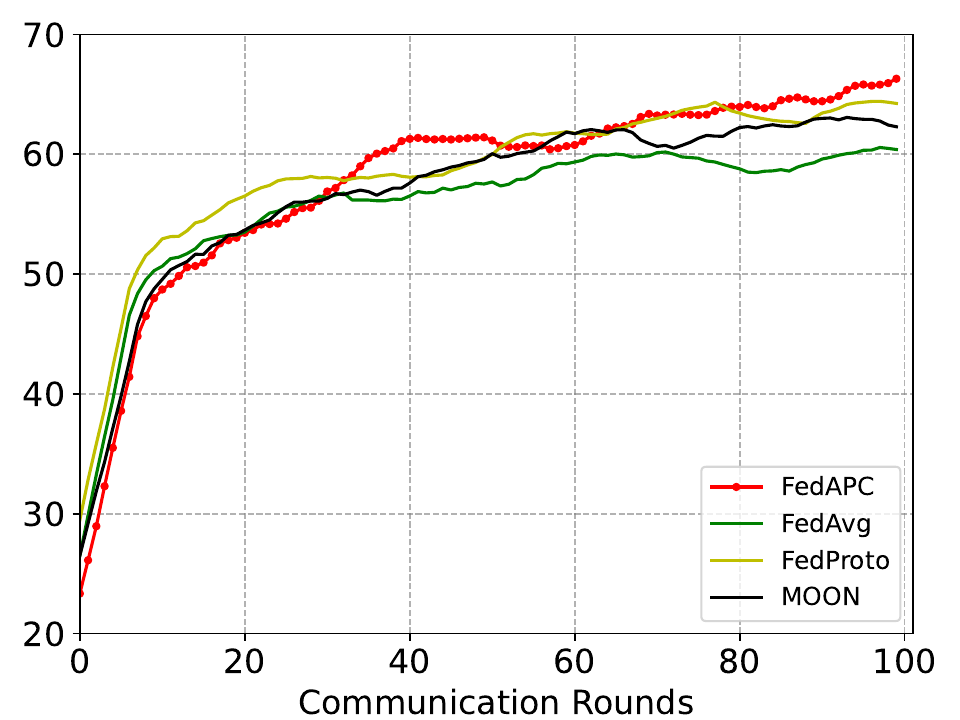}
        \caption{Office-10}
        \label{fig:dimension_hatefulmemes}
    \end{subfigure}
    \caption{Visualization of average accuracy across with $100$ communication rounds on different datasets.}
    \label{curve}
\end{figure*}

\begin{table*}[]
\caption{Comparison of performance with other SOTA methods on the all datasets. Best results are highlighted in~\textbf{bold}.}
\begin{center}
\begin{tabular}{l|ccccc|ccccc}
\toprule
\multirow{2}{*}{\textbf{Methods}} & \multicolumn{5}{c|}{Digits}                                                                                       & \multicolumn{5}{c}{Office-10}                                                                               \\
                                  & MNIST          & USPS           & SVHN          & \multicolumn{1}{c|}{SYN}            & \multicolumn{1}{l|}{Average} & Caltech        & Amazon         & Webcam         & \multicolumn{1}{c|}{DSLR}           & \multicolumn{1}{l}{Average} \\ \midrule
FedAvg        & 93.20          & \underline{84.35}          & 75.38          & \multicolumn{1}{c|}{53.91}          & 76.71                   & 63.84          & \underline{79.79}          & 53.79          & \multicolumn{1}{c|}{44.67}          & 60.52                  \\
MOON                 & 94.46          & 82.08          & \textbf{82.18}          & \multicolumn{1}{c|}{56.88}          & \underline{78.90}                    & 67.05          & 77.47          & 54.67          & \multicolumn{1}{c|}{49.33}          & 62.13                   \\
FedProto                      & \underline{94.72}          & 81.19         & 80.52          & \multicolumn{1}{c|}{\underline{58.37}}          & 78.70                    & 68.22          & \textbf{83.89}          & \underline{55.52}          & \multicolumn{1}{c|}{\underline{50.00}}          & \underline{64.41}                  \\ \midrule
\textbf{FedAPC}                                 & \textbf{95.03} & \textbf{86.51} & \underline{81.98} & \multicolumn{1}{c|}{\textbf{58.74}} & \textbf{80.57}           & \textbf{69.20}& 78.42 & \textbf{57.93}          & \multicolumn{1}{c|}{\underline{60.00}} & \textbf{66.39}          \\ 
\bottomrule
\end{tabular}
\label{comparison_digits_office}
\end{center}
\end{table*}
\begin{table}[]
\caption{Ablation study on the impact of augmentation on local prototypes in different datasets.}
\centering
\resizebox{8.5cm}{!}{%
\textcolor{black}{
\begin{tabular}{c|ccccc}
\toprule
\multicolumn{1}{c|}{\multirow{2}{*}{\begin{tabular}[c]{@{}c@{}}\textbf{Methods}\end{tabular}}} & \multicolumn{5}{c}{Digits}                                       \\
\multicolumn{1}{c|}{}                                                                                    & MNIST     & USPS     & SVHN    & \multicolumn{1}{c|}{SYN}     & Avg            \\ \midrule
\texttt{w/o} Augmentation                                                                                                & 94.26 & 84.34 & 78.43 & \multicolumn{1}{c|}{55.83} & 78.22         \\
Ours                                                                                                     & 95.03  & 86.51 & 81.98 & \multicolumn{1}{c|}{58.74} & \textbf{80.57} \\ \midrule
\multicolumn{1}{c|}{\multirow{2}{*}{\begin{tabular}[c]{@{}c@{}}\textbf{Methods}\end{tabular}}} & \multicolumn{5}{c}{Office-10}                                             \\
\multicolumn{1}{c|}{}                                                                                    & Caltech     & Amazon     & Webcam     & \multicolumn{1}{c|}{DSLR}     & Avg            \\ \midrule
\texttt{w/o} Augmentation                                                                                                & 64.17        & 78.34      & 53.86      & \multicolumn{1}{c|}{51.33}      &  61.93            \\
Ours                                                                                                     &  69.20     & 78.42      & 57.93      &  \multicolumn{1}{c|}{60.00}      & \textbf{66.39}               \\ \bottomrule
\end{tabular}}
}%
\label{tab:augmentation}
\end{table}

\section{Experiments}
\subsection{Experimental Setup}
\noindent\textbf{Datasets:}
 We assess the performance of our FedAPC using the Office-10~\cite{gong2012geodesic} and Digits~\cite{hull1994database,lecun1998gradient,netzer2011reading,roy1807effects} datasets. The Digits dataset comprises four domains: MNIST (mt), USPS (up), SVHN (sv), and SYN (syn), each containing $10$ digit classes ranging from $0$ to $9$. Similarly, Office-10 dataset consists of four domains: Caltech (C), Amazon (A), Webcam (W), and DSLR (D), also with $10$ classes. For the evaluation, we use $8$ clients and randomly distribute them across the domains as follows: Caltech: $3$, Amazon: $2$, Webcam: $2$, and DSLR: $1$ for the Office-10 dataset; and MNIST: $2$, USPS: $1$, SVHN: $2$, and SYN: $3$ for the Digits dataset. For the local training process, we sample $20\%$ of the data from the Office-10 dataset and $1\%$ from the Digits dataset for each client from the respective domains. We fix the seed to ensure the reproducibility.

\noindent\textbf{Baselines:}
 We compare our algorithm with different FL baselines, including FedAvg~\cite{mcmahan2016communication}, MOON~\cite{li2021model} and prototype-based FL method FedProto~\cite{tan2022fedproto}.

\noindent\textbf{Implementation Details:}
We implement our FedAPC algorithm using the PyTorch library~\cite{paszke2019pytorch}. For data augmentation, we apply basic techniques such as erasing, cropping, and flipping. Regarding the model architecture, we utilize ResNet-10~\cite{he2016deep}. We set the batch size to $32$, with a learning rate of $0.01$. We conduct the experiments for $3$ times and then report the average accuracy over last $5$ communication rounds.
\subsection{Experimental Results}
\noindent\textbf{Performance Comparison:}
We present a comprehensive performance comparison between our proposed FedAPC and other baselines in Table~\ref{comparison_digits_office}. The results demonstrate that FedAPC consistently surpasses the other methods, achieving superior average accuracy and accuracy across most domains. This highlights the generalization capabilities of the global model trained with FedAPC.   Notably, FedAPC achieves significantly better performance in some challenging domains, such as SYN in the Digits dataset and DSLR in the Office-10 dataset. Specifically, our method improves performance by $1.67\%$ on Digits dataset and $1.98\%$ on Office-10 dataset compared to the second-best method, respectively. These improvements underscore the robustness and adaptability of FedAPC in handling domain shifts and non-IID data distributions. Additionally, we visualize the convergence behavior across different communication rounds in Fig.~\ref{curve}. The results indicate that FedAPC not only converges faster but also exhibits more stable convergence compared to other approaches. This demonstrates that our method improves the accuracy and enhances the training stability, making it a reliable solution for federated learning in heterogeneous environments.

\noindent\textbf{Ablation study on local prototypes:}
In Table~\ref{tab:augmentation}, we evaluate the effectiveness of our proposed local augmented prototypes by comparing the performance when augmentation is removed from the local prototypes. The results clearly demonstrate that incorporating augmentation into the local prototypes leads to a noticeable improvement in performance, underscoring the significance of this approach. This observation highlights the crucial role that feature diversity plays in improving the generalization capabilities of the global model. Specifically, enhancing feature diversity through augmentation increases the robustness of the global model, enabling it to better adapt to domain heterogeneity

\section{Conclusion}
In this work, we propose FedAPC, a novel federated learning method designed to address the challenge of domain heterogeneity. By leveraging augmented prototypes and contrastive learning, FedAPC significantly enhances the robustness of the global model under domain heterogeneity, which are often induced by feature heterogeneity across distributed clients. our method effectively mitigates these challenges by aligning local feature representations with global prototypes, promoting more robust and stable model training. Experimental results on the Digits and Office-10 datasets demonstrate that FedAPC surpasses SOTA methods in both average accuracy and convergence speed. Notably, FedAPC achieves substantial improvements in challenging domains, demonstrating its ability to adapt to varied data distributions while maintaining high performance. Our approach opens the door for further advancements in federated learning, particularly in applications where domain heterogeneity is prevalent, such as in healthcare, IoT, and personalized services.

\section{Acknowledgement}
This work was supported by the Institute of Information $\&$ Communications Technology Planning $\&$ Evaluation (IITP) grant funded by the Korea government(MSIT) (No.2019-0-01287, Evolvable Deep Learning Model Generation Platform for Edge Computing),  supported by the National Research Foundation of Korea(NRF) grant funded by the Korea government(MSIT) (No. RS-2023-00207816), IITP grant funded by MSIT) (No.RS-2022-00155911, Artificial Intelligence Convergence Innovation Human Resources Development (Kyung Hee University),  IITP- ITRC(Information Technology Research Center) grant funded by MIST (IITP-2025-RS-2023-00258649), NRF grant funded by the Korea government(MSIT) (No. RS-2024-00352423), the MSIT(Ministry of Science and ICT), Korea, under the Convergence security core talent training business support program(IITP-2023(2023)- RS-2023-00266615) supervised by the IITP, and  KRIT (Korea Research Institute for defense Technology planning and advancement) grant funded by Defense Acquisition Program Administration (DAPA) (KRIT-CT-24-001). Dr. CS Hong is the corresponding author.

\bibliographystyle{IEEEtran}
\bibliography{mybib}

\end{document}